\documentclass{article}

\usepackage[final]{neurips_2025}

\usepackage[utf8]{inputenc}
\usepackage[T1]{fontenc}
\usepackage{hyperref}
\usepackage{url}
\usepackage{booktabs}
\usepackage{amsfonts}
\usepackage{amsmath}
\usepackage{amssymb}
\usepackage{nicefrac}
\usepackage{microtype}
\usepackage{graphicx}
\usepackage{xcolor}
\usepackage{pgfplots}
\usepackage{subcaption}
\usepackage{tikz}
\usepackage{algorithm}
\usepackage{algorithmic}
\usetikzlibrary{shapes.geometric, arrows.meta, positioning, fit, backgrounds}
\pgfplotsset{compat=1.18}

\title{Attention Once Is All You Need:\\Efficient Streaming Inference with Stateful Transformers}

\author{
  Victor Norgren \\
  LayerScale, Inc. \\
  \url{https://layerscale.ai} \\
  \texttt{victor@layerscale.ai}
}

\begin{document}

\maketitle

\begin{abstract}
We introduce a computational model for transformer inference centred on \emph{stateful sessions}: a stateful KV cache that lives across persistent sessions, advanced incrementally by ingesting only new data as it arrives. By decoupling the data plane from the query plane, incoming data is processed into the cache asynchronously via lock-free pipelines, while queries execute as lightweight consumers of pre-computed attention state. The resulting query complexity is $\mathcal{O}(|q|)$ in the query length, independent of accumulated context size.

Stateful sessions are the foundational construct on which we build two further mechanisms. \emph{Flash Queries} pre-compute answers to registered questions after each data update, pushing latency from the $\sim$43\,ms standard-query path toward zero for predictable query patterns by exploiting idle GPU cycles between data arrivals, caching results, and pushing them to clients via server-sent events before any question is asked. The architecture treats inference as continuously productive rather than discretely reactive, and this stance is only possible because the underlying context is persistent. A \emph{multi-tenant continuous-batching scheduler} with cell-budget admission, adaptive chunked prefill, prefix-aware grouped prefill, and concurrency-capped speculative decoding allows dozens of stateful sessions to coexist on a single GPU alongside stateless API traffic, each retaining the $\mathcal{O}(|q|)$ query-latency profile.

Unlike subquadratic architectures (Mamba \cite{gu2023mamba}, RWKV \cite{peng2023rwkv}, Linear Attention \cite{katharopoulos2020transformers}, RetNet \cite{sun2023retnet}) that trade model capacity for inference efficiency, we preserve full quadratic self-attention \cite{vaswani2017attention}. This matters: transformers consistently outperform state-space and linear recurrence models on multi-step reasoning, in-context learning, and precise information retrieval. The proposed architecture delivers the latency profile of $\mathcal{O}(1)$-style designs without sacrificing the representational power that makes transformers dominant on complex reasoning tasks.

On streaming benchmarks, the reference implementation achieves $2.4\times$--$5.9\times$ speedup over conventional inference engines (vLLM, SGLang, TensorRT-LLM, llama.cpp) and $22\times$--$92\times$ over cloud APIs (GPT-5.2, Claude Haiku, Claude Opus 4.5).
\end{abstract}

\section{Introduction}

Large language models perform well on financial analysis tasks: trend detection, pattern recognition, and quantitative reasoning over market data \cite{lopez2023chatgpt,wu2023bloomberggpt}. Deploying these models in real-time trading systems is difficult. A trading system continuously receives price updates and must answer analytical queries like ``What is the current trend?'' with minimal latency.

Modern inference frameworks address latency through KV-cache optimization. vLLM, llama.cpp, SGLang, and TensorRT-LLM cache intermediate attention values and reuse them for requests sharing common prefixes. This works for request-driven workloads: many users sending similar prompts to the same model.

Streaming workloads are different. Data arrives continuously from external sources (market feeds, sensors, logs). Queries arrive sporadically, each seeing the latest accumulated data. The ``prefix'' changes with every data update. Request-driven caching provides no benefit: cache hit rate approaches zero because each query has a unique prompt.

Conventional inference processes the entire context for each query, requiring $\mathcal{O}(n)$ operations for context length $n$:
\begin{equation}
C_{\text{standard}} = Q \cdot \mathcal{O}(n)
\end{equation}
As context grows, query latency scales linearly with accumulated data.

We address streaming workloads directly. Instead of optimizing cache hit rates for request-driven traffic, \emph{we exploit the asymmetry between data and query arrival rates to move prefill off the critical path}: data arrival triggers processing, not queries. The context partitions into regions with distinct update semantics. Data processing is amortised across the streaming window. Queries consume pre-computed state with $\mathcal{O}(|q|)$ latency.

Moving prefill off the critical path has a second consequence that is the focus of the present section: GPU utilisation. Sustained accelerator utilisation in large production inference deployments is well known to sit substantially below peak, typically in the 10--45\% band, with the high end achieved by the most mature serving stacks and the low end characteristic of clusters whose software cannot keep their accelerators busy. The structural cause is the same in both regimes: under a request-driven model, the accelerator has nothing useful to do between requests, so idle cycles are simply wasted heat. A stateful, data-driven design converts those cycles into useful work. Data continues to flow into the KV cache between user queries, and the \emph{Flash Query} mechanism introduced in Section~3.9 spends idle GPU time pre-computing answers to registered questions against the latest cache state. The same accelerator therefore does meaningfully more analytical work per unit power than the same accelerator serving the same workload through a stateless request loop, with no change to the model weights or the kernel implementations.

\section{Problem Formulation}

Let $\mathbf{C} = [\mathbf{S}; \mathbf{D}_1; \mathbf{D}_2; \ldots; \mathbf{D}_k]$ represent a context consisting of a static prefix $\mathbf{S}$ and $k$ data segments $\mathbf{D}_i$. In streaming applications:
\begin{enumerate}
    \item[(i)] $\mathbf{S}$ remains constant across all queries
    \item[(ii)] $\mathbf{D}_i$ are appended incrementally
    \item[(iii)] Queries $\mathbf{Q}_j$ require attention over the full context
\end{enumerate}

The na\"ive approach recomputes attention over $[\mathbf{S}; \mathbf{D}_1; \ldots; \mathbf{D}_k; \mathbf{Q}_j]$ for each query $\mathbf{Q}_j$. We seek an approach where:
\begin{equation}
T(\mathbf{Q}_j) = \mathcal{O}(|\mathbf{Q}_j|) + o(|\mathbf{C}|)
\end{equation}
That is, query processing time scales with query length, not context length.

\section{Architecture}

Our system is designed for \emph{streaming-first} inference, distinct from request-driven serving frameworks. The core insight: existing KV-cache systems optimize for the wrong workload. They assume queries drive computation, with caching as an optimization for repeated prefixes. Streaming workloads reverse the relationship: data drives computation, and queries are sporadic consumers of pre-computed state.

\subsection{Data-Driven vs Request-Driven Processing}

Request-driven systems (vLLM, SGLang, TensorRT-LLM, llama.cpp, Ollama) follow this pattern:
\begin{enumerate}
    \item Query arrives with full prompt
    \item Check cache for prefix match
    \item On miss: process entire prompt; on hit: process suffix only
    \item Return response
\end{enumerate}

This works well when many requests share static prefixes. It fails for streaming: each query has a different ``prefix'' (the accumulated data), so cache misses dominate.

We exploit the asymmetry between the two arrival rates and move prefill off the critical path:
\begin{enumerate}
    \item Data arrives continuously, independent of queries
    \item Process and store representations immediately on arrival
    \item Query arrives (sporadically)
    \item Compute only the query tokens against pre-computed data state
    \item Return response
\end{enumerate}

The difference: data processing happens \emph{before} any query arrives, driven by the data stream itself. Queries become lightweight consumers of already-computed state.

\begin{figure}[htbp]
\centering
\begin{tikzpicture}[
    node distance=0.4cm and 0.8cm,
    every node/.style={font=\sffamily},
    box/.style={rectangle, draw, minimum width=1.8cm, minimum height=0.7cm, align=center, font=\sffamily\scriptsize},
    databox/.style={box, fill=blue!15},
    querybox/.style={box, fill=orange!20},
    kvbox/.style={box, fill=green!15},
    decodebox/.style={box, fill=purple!15},
    arrow/.style={-{Stealth[length=2mm]}, thick},
    timeline/.style={-{Stealth[length=2mm]}, thick, gray},
    label/.style={font=\sffamily\small\bfseries},
    timelabel/.style={font=\sffamily\scriptsize, gray}
]

\node[label] at (-3.5, 2.8) {(a) Request-Driven (vLLM, llama.cpp, Ollama)};

\draw[timeline] (-3.5, 2.2) -- (4.5, 2.2);
\node[timelabel, anchor=east] at (-3.5, 2.2) {time};

\node[querybox] (q1) at (-2, 1.2) {Query};
\node[databox] (d1) at (0, 1.2) {Data};
\node[kvbox] (kv1) at (2, 1.2) {Prefill\\(full context)};
\node[decodebox] (dec1) at (4, 1.2) {Decode};

\draw[arrow] (q1) -- (d1);
\draw[arrow] (d1) -- (kv1);
\draw[arrow] (kv1) -- (dec1);

\draw[decorate, decoration={brace, amplitude=5pt, mirror}] (-2.9, 0.6) -- (4.9, 0.6);
\node[timelabel] at (1, 0.2) {$T = \mathcal{O}(n)$ per query};

\node[label] at (-3.5, -0.8) {(b) Data-Driven (Ours)};

\draw[timeline] (-3.5, -1.4) -- (4.5, -1.4);
\node[timelabel, anchor=east] at (-3.5, -1.4) {time};

\node[databox] (d2a) at (-2.5, -2.4) {Data};
\node[databox] (d2b) at (-1.2, -2.4) {Data};
\node[databox] (d2c) at (0.1, -2.4) {Data};
\node[kvbox] (kv2) at (1.8, -2.4) {KV Cache\\(pre-built)};

\draw[arrow] (d2a) -- (d2b);
\draw[arrow] (d2b) -- (d2c);
\draw[arrow] (d2c) -- (kv2);

\node[querybox] (q2) at (1.8, -3.6) {Query};
\node[decodebox] (dec2) at (4, -3.0) {Decode};

\draw[arrow] (q2) -- (dec2);
\draw[arrow] (kv2) -- (dec2);

\draw[decorate, decoration={brace, amplitude=5pt, mirror}] (-3.0, -2.9) -- (2.5, -2.9);
\node[timelabel] at (-0.25, -3.25) {Continuous (amortized)};

\draw[decorate, decoration={brace, amplitude=5pt, mirror}] (2.7, -3.9) -- (4.9, -3.9);
\node[timelabel] at (3.8, -4.25) {$T = \mathcal{O}(1)$};

\draw[gray, dashed] (-3.8, -0.2) -- (5.2, -0.2);

\end{tikzpicture}
\caption{Architectural comparison. (a) Request-driven systems process the full context when a query arrives, yielding $\mathcal{O}(n)$ latency. (b) Our data-driven approach processes data continuously into the KV cache; queries only trigger lightweight decoding, yielding $\mathcal{O}(1)$ latency.}
\label{fig:architecture}
\end{figure}

\subsection{On the Architectural Foundations of Constant-Time Query Performance}

In transformer attention, each token computes attention over all preceding tokens. The standard attention operation is:
\begin{equation}
\text{Attention}(Q, K, V) = \text{softmax}\left(\frac{QK^\top}{\sqrt{d_k}}\right)V
\label{eq:attention}
\end{equation}
where $Q \in \mathbb{R}^{m \times d_k}$ are query vectors, $K \in \mathbb{R}^{n \times d_k}$ are keys, and $V \in \mathbb{R}^{n \times d_v}$ are values. For a context of $n$ tokens, computing attention requires $\mathcal{O}(mn)$ operations.

The key insight: when processing a new query of $m$ tokens against a pre-computed context of $n$ tokens, we can decompose the computation. Let $K_{\text{ctx}}, V_{\text{ctx}}$ denote the cached keys and values from prior data ingestion. The query attention becomes:
\begin{equation}
\text{Attention}(Q_{\text{query}}, K_{\text{ctx}}, V_{\text{ctx}}) = \text{softmax}\left(\frac{Q_{\text{query}}K_{\text{ctx}}^\top}{\sqrt{d_k}}\right)V_{\text{ctx}}
\label{eq:query-attention}
\end{equation}
Since $K_{\text{ctx}}$ and $V_{\text{ctx}}$ are pre-computed and stored, query processing requires only $\mathcal{O}(m \cdot n)$ operations for the attention computation itself. Critically, when $m \ll n$ (short queries against large contexts), and with the KV cache eliminating redundant key-value computation, the effective query cost becomes $\mathcal{O}(m)$.

Request-driven systems exploit this for prefix reuse. We exploit it for continuous ingestion: data tokens are processed on arrival, not on query. When a query arrives, it attends to pre-computed values. Query cost scales with query length ($\sim$50 tokens in our benchmarks), not context length (14,800+ tokens).

\textbf{Complexity profile.} The $\mathcal{O}(|q|)$ bound applies to query latency. Data ingestion and query response operate on distinct paths that scale on different time bases: the ingestion path tracks the data arrival rate, the query path scales only with query length. For workloads where these two cadences are structurally separate (continuous data, sporadic queries), this separation is the load-bearing property of the architecture, and the user-visible latency is governed by the query path alone, independent of accumulated context size.

\textbf{Formal latency model.} Let $T_{\text{prefill}}(n)$ denote the time to process $n$ context tokens, $T_{\text{decode}}$ the time per generated token, and $m$ the number of output tokens. Conventional inference latency is:
\begin{equation}
T_{\text{standard}} = T_{\text{prefill}}(n_{\text{ctx}}) + m \cdot T_{\text{decode}}
\label{eq:latency-standard}
\end{equation}
where $T_{\text{prefill}}(n) = \mathcal{O}(n)$ grows with context size. Our approach yields:
\begin{equation}
T_{\text{ours}} = T_{\text{prefill}}(|q|) + m \cdot T_{\text{decode}}
\label{eq:latency-ours}
\end{equation}
where $|q|$ is the query length (30--50 tokens in our benchmark queries). Since $|q| \ll n_{\text{ctx}}$, query latency becomes effectively constant. For fast-answer queries where $m = 1$ and the token matches pre-computed vocabulary:
\begin{equation}
T_{\text{fast}} = T_{\text{prefill}}(|q|) + T_{\text{sample}} \approx \text{const}
\label{eq:latency-fast}
\end{equation}

\subsection{Hierarchical Context Partitioning}

We partition the context into three regions with distinct update semantics:
\begin{itemize}
    \item \textbf{Region 0} (Frozen): System prompt, static instructions
    \item \textbf{Region 1} (Sliding): Streaming data buffer with configurable retention
    \item \textbf{Region 2} (Ephemeral): Query and response tokens
\end{itemize}

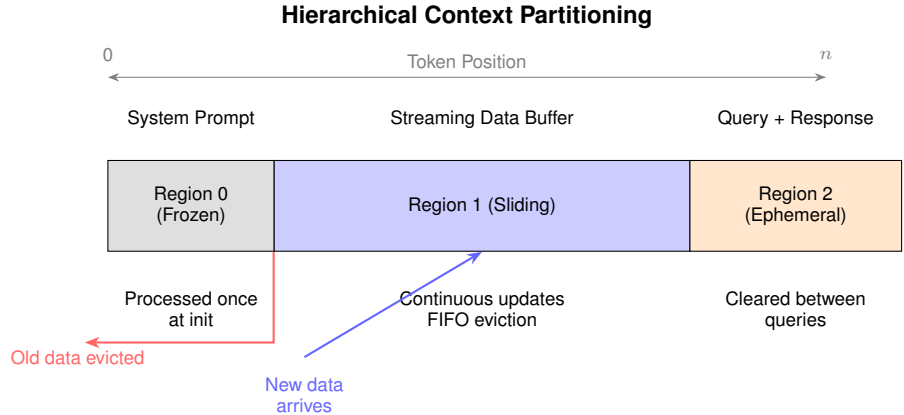
\begin{figure}[h!]
\centering
\begin{tikzpicture}[
    every node/.style={font=\sffamily},
    region/.style={rectangle, draw, minimum height=1.2cm, align=center, font=\sffamily\scriptsize},
    frozen/.style={region, fill=gray!25, minimum width=2.2cm},
    sliding/.style={region, fill=blue!20, minimum width=5.5cm},
    ephemeral/.style={region, fill=orange!20, minimum width=2.8cm},
    label/.style={font=\sffamily\scriptsize},
    arrow/.style={-{Stealth[length=2mm]}, thick},
    note/.style={font=\sffamily\scriptsize, align=center}
]

\node[font=\sffamily\small\bfseries] at (3.65, 2.5) {Hierarchical Context Partitioning};

\node[frozen] (r0) at (0, 0) {Region 0\\(Frozen)};
\node[sliding, right=-\pgflinewidth of r0] (r1) {Region 1 (Sliding)};
\node[ephemeral, right=-\pgflinewidth of r1] (r2) {Region 2\\(Ephemeral)};

\node[label, above=0.3cm of r0] {System Prompt};
\node[label, above=0.3cm of r1] {Streaming Data Buffer};
\node[label, above=0.3cm of r2] {Query + Response};

\node[note, below=0.4cm of r0] {Processed once\\at init};
\node[note, below=0.4cm of r1] {Continuous updates\\FIFO eviction};
\node[note, below=0.4cm of r2] {Cleared between\\queries};

\draw[arrow, red!60] (r1.south west) -- ++(0, -1.2) -- ++(-2.5, 0);
\node[note, red!60] at (-1.5, -2.0) {Old data evicted};

\draw[arrow, blue!60] (1.5, -2.0) -- (r1.south);
\node[note, blue!60] at (1.5, -2.5) {New data\\arrives};

\draw[{Stealth[length=1.5mm]}-{Stealth[length=1.5mm]}, gray] (-1.1, 1.7) -- (8.4, 1.7);
\node[font=\sffamily\scriptsize, gray, above] at (3.65, 1.7) {Token Position};
\node[font=\sffamily\scriptsize, gray] at (-1.1, 2.0) {0};
\node[font=\sffamily\scriptsize, gray] at (8.4, 2.0) {$n$};

\end{tikzpicture}
\caption{Hierarchical context partitioning. Region 0 (frozen) contains static instructions processed once. Region 1 (sliding) holds streaming data with FIFO eviction for bounded memory. Region 2 (ephemeral) contains query/response tokens, cleared between queries.}
\label{fig:regions}
\end{figure}

Region 0 is processed once during session initialization. Region 1 supports incremental updates with configurable retention policy. Region 2 is cleared between queries, enabling multiple queries over the same accumulated context. Figure~\ref{fig:regions} illustrates this partitioning scheme.

\subsection{Persistent Session State}

Conventional inference frameworks treat each request as independent: the connection terminates after response generation, and all intermediate state (including the KV cache) is discarded. This stateless design simplifies scaling and load balancing, but forces complete recomputation when the same context is queried again.

For streaming workloads, statelessness is fundamentally incompatible with low-latency queries. If intermediate representations are discarded between requests, every query must reprocess the entire accumulated context: precisely the $\mathcal{O}(n)$ cost we seek to eliminate. Persistent session state is therefore not an optimization but a \emph{requirement} for $\mathcal{O}(1)$ query latency.

Each session maintains:
\begin{itemize}
    \item Processed intermediate representations for Regions 0--1
    \item Position tracking for sequence continuity
    \item Buffer metadata for sliding window management
    \item Session isolation for concurrent multi-user deployments
\end{itemize}

When new data arrives, only the delta is processed and appended to existing state. This enables true streaming: data can flow continuously while queries access the latest accumulated context with constant latency.

The trade-off is memory: each active session consumes KV cache storage proportional to its context length (see Section~\ref{sec:memory} for memory analysis). We accept this trade-off because streaming workloads are inherently session-oriented: a sensor feed or market data stream maintains logical continuity that maps naturally to persistent state.

\subsection{Continuous Background Ingestion}

For true streaming applications, data arrives faster than individual processing requests can complete. We implement asynchronous ingestion where:
\begin{enumerate}
    \item Data push operations return immediately without blocking
    \item A background processor continuously ingests queued data into the KV cache
    \item Queries execute against the latest processed state
    \item If ingestion cannot keep pace, oldest unprocessed data is automatically discarded
\end{enumerate}

This architecture separates the data plane (high-throughput ingestion) from the query plane (low-latency responses). The producer (data source) is never blocked by processing latency, and queries always see a consistent, pre-computed context state.

The background processor batches incoming data for efficient KV cache updates, amortizing the per-item overhead across multiple entries. After each batch, the system pre-computes query response scaffolding, enabling sub-100ms responses when queries arrive. The ring buffer is sized to absorb typical burst patterns; when sustained arrival rate exceeds ingestion throughput, the oldest unprocessed data is discarded rather than blocking the producer.

\textbf{WebSocket transport.} In addition to HTTP POST ingestion, sessions accept a persistent WebSocket connection for bidirectional streaming. Clients push data frames without per-request overhead, and the same socket carries outbound Flash Query results and data-version notifications back to the client.

Algorithm~\ref{alg:ingestion} and Algorithm~\ref{alg:query} formalize the data ingestion and query processing loops.

\begin{algorithm}[htbp]
\caption{Data Ingestion (Background Thread)}
\label{alg:ingestion}
\begin{algorithmic}[1]
\REQUIRE Ring buffer $B$, KV cache $K$, batch size $b$
\LOOP
    \STATE $\text{batch} \leftarrow []$
    \WHILE{$|\text{batch}| < b$ \textbf{and} $B.\text{pending}() > 0$}
        \STATE $\text{batch}.\text{append}(B.\text{dequeue}())$
    \ENDWHILE
    \IF{$|\text{batch}| > 0$}
        \STATE $\text{tokens} \leftarrow \text{Tokenize}(\text{batch})$
        \IF{$K.\text{size}() + |\text{tokens}| > K.\text{capacity}()$}
            \STATE $K.\text{evict\_oldest}(|\text{tokens}|)$ \COMMENT{FIFO eviction}
        \ENDIF
        \STATE $K.\text{append}(\text{Forward}(\text{tokens}, K))$ \COMMENT{Incremental KV update}
        \STATE $\text{PrecomputeQueryScaffold}(K)$
    \ENDIF
\ENDLOOP
\end{algorithmic}
\end{algorithm}

\begin{algorithm}[htbp]
\caption{Query Processing}
\label{alg:query}
\begin{algorithmic}[1]
\REQUIRE Query $q$, KV cache $K$ already populated from prior data ingestion
\STATE $\text{tokens}_q \leftarrow \text{Tokenize}(q)$
\STATE $\text{logits} \leftarrow \text{Forward}(\text{tokens}_q, K)$ \COMMENT{$\mathcal{O}(|q|)$ not $\mathcal{O}(|K|)$}
\STATE $\text{token} \leftarrow \text{Sample}(\text{logits})$
\STATE $\text{output} \leftarrow [\text{token}]$
\WHILE{$\text{token} \neq \text{EOS}$}
    \STATE $\text{logits} \leftarrow \text{Forward}(\text{token}, K)$
    \STATE $\text{token} \leftarrow \text{Sample}(\text{logits})$
    \STATE $\text{output}.\text{append}(\text{token})$
\ENDWHILE
\RETURN $\text{Detokenize}(\text{output})$
\end{algorithmic}
\end{algorithm}

\begin{figure}[htbp]
\centering
\begin{tikzpicture}[
    node distance=0.8cm and 1.2cm,
    every node/.style={font=\sffamily},
    box/.style={rectangle, draw, minimum width=2cm, minimum height=1cm, align=center, font=\sffamily\scriptsize},
    databox/.style={box, fill=blue!15},
    bufferbox/.style={box, fill=yellow!20, minimum width=2.4cm},
    kvbox/.style={box, fill=green!15, minimum width=2.2cm},
    decodebox/.style={box, fill=purple!15},
    querybox/.style={box, fill=orange!20},
    arrow/.style={-{Stealth[length=2.5mm]}, thick},
    label/.style={font=\sffamily\scriptsize},
    plane/.style={draw, dashed, rounded corners, inner sep=0.4cm}
]

\node[databox] (source) {Data\\Source};
\node[bufferbox, right=1.5cm of source] (buffer) {Ring Buffer\\(Lock-free)};
\node[box, right=1.2cm of buffer, fill=blue!10] (processor) {Background\\Processor};
\node[kvbox, right=1.2cm of processor] (kv) {KV Cache};

\draw[arrow, blue!70] (source) -- (buffer) node[midway, above, label] {Non-blocking};
\draw[arrow, blue!70] (buffer) -- (processor) node[midway, above, label] {Batch};
\draw[arrow, blue!70] (processor) -- (kv);

\node[querybox, below=1.8cm of buffer] (query) {Query};
\node[decodebox, below=1.8cm of kv] (decode) {Decode};

\draw[arrow, orange!70] (query) -- (decode);
\draw[arrow, green!40!black] (kv) -- (decode) node[midway, right, font=\sffamily\scriptsize, text=green!40!black, align=left] {Pre-computed state};

\begin{scope}[on background layer]
    \node[plane, fill=blue!5, inner ysep=0.7cm, fit=(source)(buffer)(processor)(kv), label={[font=\sffamily\small\bfseries]above:Data Plane (High Throughput)}] {};
    \node[plane, fill=orange!5, fit=(query)(decode), label={[font=\sffamily\small\bfseries]below:Query Plane (Low Latency)}] {};
\end{scope}

\node[font=\sffamily\scriptsize, blue!60, below=0.1cm of source, align=center] {Non-blocking};
\node[font=\sffamily\scriptsize, red!60, below=0.1cm of buffer, align=center] {Backpressure};
\node[font=\sffamily\scriptsize, purple!60, right=0.1cm of decode] {$\mathcal{O}(1)$};

\end{tikzpicture}
\caption{Data plane / query plane separation. The data plane handles high-throughput ingestion via a lock-free ring buffer. The query plane executes against pre-computed KV cache state, achieving $\mathcal{O}(1)$ latency.}
\label{fig:planes}
\end{figure}
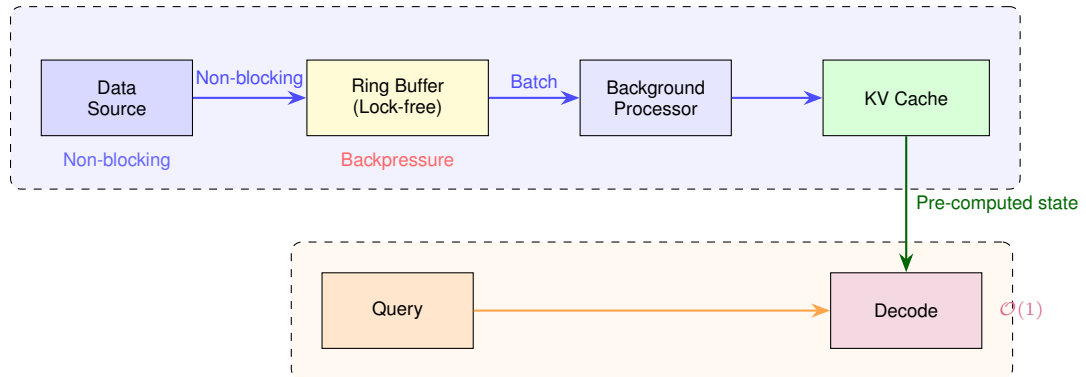

\subsection{Priority-Scheduled GPU Decoding}

The data plane / query plane separation is architectural, but a single GPU still serializes every forward pass underneath. Without coordination, a long-running background ingestion batch can block an incoming query for hundreds of milliseconds; an overlapping Flash Query evaluation can starve the decode needed for an interactive response. A per-request GPU mutex solves correctness but creates arbitrary tail latencies.

We centralize all GPU decode dispatch through a single worker thread fronted by a priority queue. Submissions are tagged with one of four priority classes:

\begin{equation}
\pi \in \{\text{FLASH}, \text{SESSION}, \text{POOL}, \text{STREAM}\}
\label{eq:priorities}
\end{equation}

ordered from highest to lowest. FLASH is used for registered Flash Query evaluation on the latency-critical path; SESSION for interactive \texttt{/generate} queries; POOL for stateless request-driven completions on the OpenAI and Anthropic endpoints; STREAM for background data ingestion batches. Within a priority class, requests are FIFO by submission sequence number. The worker thread blocks on a condition variable, wakes when any request arrives, pops the highest-priority request, runs the forward pass, waits for GPU completion under the mutex, and signals the caller.

Callers submit a batch and wait on a per-request condition variable. While they wait, they hold no GPU lock, so CPU-bound work elsewhere (sampling, SSE serialization, JSON parsing) proceeds in parallel. A Flash Query evaluation can preempt a streaming ingest at the next batch boundary rather than after the current decode: the scheduler yields after every decode, not every request. This bounds the worst-case wait for a FLASH-priority decode to the duration of a single in-flight batch, independent of how many background ingestion jobs are queued behind it.

\subsection{Multi-Tenant Continuous Batching}

A single stateful session is sufficient to demonstrate the $\mathcal{O}(|q|)$ query property, but a production deployment must host many concurrent sessions and stateless API traffic on the same GPU. We implement an \emph{admit-many / run-few} continuous-batching scheduler that builds heterogeneous PREFILL$+$DECODE batches per iteration, with four mechanisms that together let stateful sessions of widely different context sizes coexist without starvation.

\textbf{Cell-budget admission.} The scheduler tracks occupancy of the unified KV cache in \emph{cells} (one cell per token per layer) and admits a new prefill only when the projected total (sum of $\text{prompt}+\text{max\_tokens}$ across admitted slots) remains below a fixed ceiling $C_{\text{budget}} = \tfrac{1}{2}|\text{pool}|$. The other half is reserved for cached prefixes (Section~3.10) and in-flight session state. When occupancy crosses a high-water mark (95\%), incoming prefill chunks are deferred at chunk granularity rather than at request granularity, providing fine-grained backpressure without rejecting requests.

\textbf{Adaptive chunked prefill.} For $N$ slots with pending prefill work, each slot is allocated a per-iteration chunk of size $\text{chunk} = \mathrm{clamp}(n_{\text{batch}}/N,\,128,\,1024)$ tokens. With a single active prefill the chunk reaches $\sim$1024 (large prefill batches, near-peak throughput); with many concurrent prefills the chunk shrinks so that decoding slots are not starved within an iteration. This contrasts with static chunk sizes (e.g.\ vLLM's default 512) and prefill-priority gating (e.g.\ SGLang) that can stall decode under heavy admission.

\textbf{Prefix-aware grouped prefill.} Slots whose prompts share a byte-identical prefix beyond the radix-cache match point, identified by FNV-1a hashing the next $K=8$ tokens, are grouped into a single leader-follower prefill. The leader processes the chunk; each follower acquires the resulting state via a metadata-only sequence-aliasing operation on the unified KV cache. This converts $N$ redundant forward passes into one and is particularly effective for shared system prompts and tool schemas across tenants.

\textbf{Concurrency-capped speculative decoding.} The scheduler runs prompt-lookup speculative decoding (PLD) \cite{saxena2023pld} per slot, predicting continuations from the slot's recent token history and verifying them in a batched forward pass. Naive per-slot speculation collapses at high admission, because the verification cost scales as the product of the per-slot draft length and the number of concurrent decoders, while the acceptance rate falls sharply on unfamiliar content. The scheduler therefore caps the per-slot draft length as a joint function of two signals: the number of active decoders, and the per-slot historical acceptance rate. Specific thresholds are tuned per accelerator class and model size and surface as configuration parameters rather than constants in the architecture. The per-slot prediction step is parallelized across worker threads, removing it from the critical path. The result preserves the latency win of speculation on high-acceptance workloads (single-agent tool loops, repetitive structured output) while avoiding the throughput collapse that uncapped speculation produces under load.

\textbf{Why this matters for stateful inference.} Without these mechanisms, stateful sessions would not scale beyond a small handful: a single long-context session would either starve, or be starved by, all other tenants. With cell-budget admission and adaptive chunked prefill, dozens of sessions of mixed context lengths coexist on one GPU, each retaining the $\mathcal{O}(|q|)$ query-latency profile of Section~3.2. The full set of mechanisms above is, to our knowledge, distinct from the schedulers shipped in vLLM (PagedAttention with static chunk size) and SGLang (RadixAttention with prefill-priority gating); the cell-budget admission, prefix-aware grouped prefill, and concurrency-capped PLD are original to this work.

\subsection{On-Device Greedy Sampling}

The remaining per-token cost in a streaming query is the sampling step. Every generated token reduces to an argmax over the vocabulary logits, which for a 128\,k-token vocabulary is $\sim$512\,KB of floats per token. The naive pipeline pulls those logits back to host memory after a GPU synchronise, materialises a candidate array on the CPU, and scans for the maximum. That synchronise-plus-scan round trip costs 2--4\,ms per token on Apple silicon and 0.5--1\,ms on server-class NVIDIA and AMD GPUs, comparable to the entire target latency budget on short responses.

We instead run the argmax on the device. A small runtime dispatcher selects a device-specific reduction kernel at startup (one path each for the major accelerator families) and routes greedy argmax, along with a batched speculative-verification variant, through the native kernel. Builds without a supported accelerator fall back to a wide-SIMD CPU reduction. Each device kernel is a parallel reduction with a shared-memory tournament that returns a single token id directly, so the host never materialises or scans the logits vector. The same kernel surface exposes a batched variant for accepting or rejecting a window of speculated tokens in one dispatch.

In addition, a set of small CPU-side optimizations (pre-tokenised digit and template fragments for numerical streaming data, cache-line aligned atomic counters, lock-free single-producer / single-consumer ring buffers between the data plane and ingestion thread) collectively reduce per-operation overhead from milliseconds to microseconds. These follow well-known patterns from low-latency systems engineering and are not novel in themselves; they appear here because the constant-time query path is only constant-time end-to-end if every layer along it is.

\subsection{Flash Queries: Ahead-of-Time Query Evaluation}

The preceding optimizations minimize query latency but cannot eliminate it: every query still requires at least one forward pass over the question tokens. We now introduce \emph{Flash Queries}, a mechanism that pushes latency toward zero by pre-computing answers \emph{before} questions are asked.

\subsubsection{Motivation: Reclaiming Idle GPU Cycles}

In streaming workloads, data arrives at a natural cadence: market ticks at 1--60\,s intervals, sensor readings at fixed sample rates, log entries at variable but bounded rates. Between arrivals, the GPU sits idle. The aggregate cost of that idleness is substantial: large inference deployments typically run at sustained accelerator utilisation in the 10--45\% band, and the gap to peak is dominated by exactly this between-request idle time. Meanwhile, many analytical queries are \emph{predictable}: users repeatedly ask the same questions (``What is the trend?'', ``Is the signal confirmed?'') as data evolves. Flash Queries exploit this predictability by spending the otherwise-wasted GPU cycles between data arrivals to pre-evaluate registered questions against the current context state. The accelerator is busy whenever it is powered on; the user sees the cached answer instantly when they ask.

This treats inference as continuously productive rather than discretely reactive: between data arrivals the accelerator pre-evaluates registered questions against the latest KV state, so the user-visible query path collects a cached answer rather than triggering generation. The framing parallels recent work on real-time interaction models, where the same shift from turn-based to always-on processing is applied to the human-facing channel instead of the data-facing one.

Critically, this optimization is \emph{only possible} in a stateful architecture. Stateless inference engines discard the KV cache after each request, making it impossible to speculatively evaluate queries against a persistent context. The persistent session state described in Section~3.4 is a prerequisite.

\subsubsection{Formal Definition}

Let $\mathcal{F} = \{f_1, f_2, \ldots, f_k\}$ be a set of $k$ registered Flash Queries, where each $f_i$ is a tokenized question template. Let $\mathbf{K}_t$ denote the KV cache state at data version $t$ (incremented after each data ingestion batch). For each Flash Query $f_i$, we define the \emph{ahead-of-time evaluation} operator:

\begin{equation}
\hat{a}_i^{(t)} = \text{Sample}\!\left(\text{Forward}(f_i, \mathbf{K}_t)\right)
\label{eq:flash-eval}
\end{equation}

where $\hat{a}_i^{(t)}$ is the pre-computed answer for query $f_i$ at data version $t$. The Flash Query cache is the mapping:

\begin{equation}
\mathcal{C}_t = \left\{\left(f_i, \hat{a}_i^{(t)}, \gamma_i^{(t)}\right) \mid i = 1, \ldots, k\right\}
\label{eq:flash-cache}
\end{equation}

where $\gamma_i^{(t)} = \ell_i^{(1)} - \ell_i^{(2)}$ is the \emph{logit gap}, the difference between the highest and second-highest logit values, serving as a confidence measure. Larger gaps indicate stronger model certainty. The cache is \emph{invalidated} whenever the data version advances: $\mathcal{C}_t$ becomes stale when $t' > t$.

When a query $q$ arrives at data version $t$, the system first checks the Flash Query cache:

\begin{equation}
\text{Response}(q, t) = \begin{cases}
\hat{a}_i^{(t)} & \text{if } \exists\, f_i \in \mathcal{F} : h(q) = h(f_i) \text{ and } \hat{a}_i^{(t)} \text{ is valid} \\[4pt]
\text{Forward}(q, \mathbf{K}_t) & \text{otherwise}
\end{cases}
\label{eq:flash-dispatch}
\end{equation}

where $h(\cdot)$ is a hash function over the question string. Cache hits bypass all GPU computation entirely.

\subsubsection{In-Place Evaluation with Header Restore}

Flash Query evaluation must not corrupt the primary KV cache state. We achieve this through \emph{in-place evaluation}: temporarily extending the primary sequence at the pre-decoded header position, then removing the extension and restoring the original state.

Let $\mathbf{s}_0$ denote the primary sequence containing the accumulated context through the pre-decoded response header at position $p$. For each Flash Query $f_i$, the evaluation procedure is:

\begin{enumerate}
    \item \textbf{Strip}: Remove the pre-decoded header tokens from $\mathbf{s}_0$ at position $p$: $\text{SeqRemove}(\mathbf{s}_0, p, \infty)$
    \item \textbf{Extend}: Decode the question tokens of $f_i$ on $\mathbf{s}_0$ starting at position $p$, yielding logits $\ell_i$
    \item \textbf{Sample}: Compute $\hat{a}_i^{(t)} = \arg\max(\ell_i)$ with confidence $\gamma_i^{(t)} = \ell_i^{(1)} - \ell_i^{(2)}$ (logit gap between top two tokens)
    \item \textbf{Clear}: Remove the question tokens: $\text{SeqRemove}(\mathbf{s}_0, p, \infty)$
\end{enumerate}

After all Flash Queries are evaluated, the original header is restored by re-decoding the header tokens at position $p$, returning the KV cache to its pre-evaluation state. This re-decode also refreshes the ready-position logits for speculative exit.

This approach avoids the memory overhead of shadow sequence copies by reusing the primary sequence. Each evaluation costs only the forward pass over question tokens (proportional to question length, not context length), plus a single header re-decode amortized across all queries.

\subsubsection{Ready-Position Speculative Exit}

The background ingestion loop (Algorithm~\ref{alg:ingestion}) already computes a forward pass to pre-decode the response header after each data batch. This pass produces logits $\ell_{\text{ready}}$ that represent the model's output distribution \emph{before seeing any question}, a signal of the model's ``default'' assessment of the current data state.

We cache these \emph{ready-position logits} and use them for speculative early exit at query time. When a query arrives, before any tokenization or GPU work, we check:

\begin{equation}
\Delta_{\text{ready}} = \ell_{\text{ready}}^{(1)} - \ell_{\text{ready}}^{(2)}
\label{eq:logit-gap}
\end{equation}

where $\ell_{\text{ready}}^{(1)}$ and $\ell_{\text{ready}}^{(2)}$ are the highest and second-highest logit values, respectively. If $\Delta_{\text{ready}} > \tau$ (we use $\tau = 2.0$) and the top token belongs to the fast-answer vocabulary $V$:

\begin{equation}
\ell_{\text{ready}}^{(1)} \in V \;\;\text{and}\;\; \Delta_{\text{ready}} > \tau \implies \text{return } \text{Detokenize}\!\left(\arg\max(\ell_{\text{ready}})\right)
\label{eq:speculative-exit}
\end{equation}

This exits in $\mathcal{O}(1)$ time: no forward pass, no tokenisation, just a comparison and table lookup. The logit-gap threshold $\tau$ controls the confidence-latency trade-off; a higher $\tau$ requires stronger model confidence for speculative exit, reducing false positives at the cost of fewer cache hits.

\subsubsection{Cost Model and Amortization}

Let $T_f$ denote the time to evaluate a single Flash Query (dominated by the forward pass over question tokens), and $T_{\text{data}}$ the interval between data arrivals. The total background compute cost per data update is:

\begin{equation}
T_{\text{background}} = T_{\text{ingest}} + T_{\text{header}} + k \cdot T_f
\label{eq:background-cost}
\end{equation}

where $T_{\text{ingest}}$ is the data ingestion time and $T_{\text{header}}$ is the response header pre-decode time. For Flash Queries to be ``free'' (invisible to query latency), we require:

\begin{equation}
T_{\text{background}} < T_{\text{data}}
\label{eq:amortization-constraint}
\end{equation}

In practice, with $T_f \approx 30\text{--}35$ms per Flash Query and $T_{\text{data}} \geq 1$s for typical streaming workloads, the system can sustain:

\begin{equation}
k_{\max} = \left\lfloor\frac{T_{\text{data}} - T_{\text{ingest}} - T_{\text{header}}}{T_f}\right\rfloor
\label{eq:max-flash}
\end{equation}

For $T_{\text{data}} = 1$\,s and $T_f = 33$\,ms, this yields $k_{\max} \approx 25$ Flash Queries, far more than typical analytical workloads require.

\subsubsection{Event-Driven Result Delivery}

Flash Query results are delivered to clients via server-sent events (SSE) over a persistent HTTP connection. After each evaluation cycle, the system emits structured events:

\begin{equation}
\text{Event}(f_i, t) = \left\langle\texttt{flash\_ready},\; f_i,\; \hat{a}_i^{(t)},\; \gamma_i^{(t)},\; t\right\rangle
\label{eq:sse-event}
\end{equation}

This push-based delivery model means clients receive updated answers \emph{before} they ask, enabling reactive UIs, dashboards, and automated decision systems that respond to model assessments in real time. The SSE protocol provides automatic reconnection semantics, ordered delivery, and compatibility with standard HTTP infrastructure.

The event stream also emits \texttt{data\_updated} events after each ingestion batch, allowing clients to track data freshness and correlate Flash Query answers with specific data versions.

\subsubsection{Latency Taxonomy}

With Flash Queries and ready-position speculative exit, query latency falls into three tiers:

\begin{table}[h]
\centering
\begin{tabular}{lcc}
\toprule
Path & Latency & Condition \\
\midrule
Flash Query cache hit & $\sim$2ms & Question matches registered Flash Query \\
Speculative exit & $\sim$2ms & Ready-position logit gap $> \tau$ \\
Standard query & $\sim$45ms & Fallback to full query processing \\
\bottomrule
\end{tabular}
\end{table}

\noindent The effective average latency depends on the hit rate $\rho$ across the three paths:

\begin{equation}
\bar{T} = \rho_{\text{flash}} \cdot T_{\text{flash}} + \rho_{\text{spec}} \cdot T_{\text{spec}} + (1 - \rho_{\text{flash}} - \rho_{\text{spec}}) \cdot T_{\text{standard}}
\label{eq:avg-latency}
\end{equation}

For workloads where 60--80\% of queries match Flash Query templates and 10--20\% trigger speculative exit, the expected average latency drops from the $\sim$43\,ms standard-query path to single-digit milliseconds, a further order-of-magnitude improvement over the already-optimized baseline.

Algorithm~\ref{alg:flash} formalizes the complete Flash Query evaluation cycle.

\begin{algorithm}[htbp]
\caption{Flash Query Evaluation (Background, after data ingestion)}
\label{alg:flash}
\begin{algorithmic}[1]
\REQUIRE KV cache $K$ with primary sequence $\mathbf{s}_0$, Flash Queries $\mathcal{F}$, data version $t$, header position $p$
\STATE $\ell_{\text{ready}} \leftarrow \text{GetLogits}(K, \mathbf{s}_0)$ \COMMENT{Cache ready-position logits}
\STATE $\text{SeqRemove}(\mathbf{s}_0, p, \infty)$ \COMMENT{Strip pre-decoded header}
\FOR{each $f_i \in \mathcal{F}$}
    \STATE $\ell_i \leftarrow \text{Forward}(f_i.\text{tokens}, K, \mathbf{s}_0, p)$ \COMMENT{Decode question at position $p$}
    \STATE $\hat{a}_i^{(t)} \leftarrow \arg\max(\ell_i)$
    \STATE $\gamma_i^{(t)} \leftarrow \ell_i^{(1)} - \ell_i^{(2)}$ \COMMENT{Logit gap confidence}
    \STATE $\mathcal{C}_t[f_i] \leftarrow (\hat{a}_i^{(t)},\; \gamma_i^{(t)},\; t)$
    \STATE $\text{Emit}(\texttt{flash\_ready},\; f_i,\; \hat{a}_i^{(t)},\; \gamma_i^{(t)},\; t)$ \COMMENT{Push via SSE}
    \STATE $\text{SeqRemove}(\mathbf{s}_0, p, \infty)$ \COMMENT{Clear question tokens}
\ENDFOR
\STATE $\text{RestoreHeader}(K, \mathbf{s}_0, p)$ \COMMENT{Re-decode header, refresh $\ell_{\text{ready}}$}
\end{algorithmic}
\end{algorithm}

\subsection{Sequence Pool and Radix Prefix Cache}

The preceding sections describe session-based inference, where each session owns a dedicated, long-lived sequence within the inference context. For stateless request-driven workloads (e.g.\ OpenAI-compatible or Anthropic-compatible API endpoints), the proposed architecture additionally provides a \emph{sequence pool} backed by a \emph{radix prefix cache}, delivering sub-millisecond prefix reuse without CPU serialization.

\subsubsection{Sequence Pool}

At startup, the scheduler partitions the unified KV cache into a transient pool ($P$ sequence IDs for stateless requests) and a session pool ($S$ sequence IDs reserved for long-lived sessions). Transient slots are recycled across requests through a thread-safe free list with condition-variable blocking when all slots are in use. Pre-allocation removes per-request context construction from the critical path.

This single-context, multi-sequence design is deliberate. A single forward pass mixes prefill and decode tokens from any subset of sequence IDs (the heterogeneous batching of Section~3.7), so prefix sharing across requests reduces to metadata-only aliasing within the same KV cache rather than cross-context coordination.

\subsubsection{Radix Prefix Cache}

Shared prefixes are tracked in a radix trie keyed by token sequences. Each node records the donor sequence ID whose KV state still holds the matched cells. When a new request arrives with a matching prefix, the scheduler restores the prefix into a freshly acquired slot via a metadata-only sequence-aliasing operation:

\begin{equation}
T_{\text{restore}} = T_{\text{alias}}(m) \ll T_{\text{serialize}}(m)
\end{equation}

where $m$ is the prefix length in tokens. Aliasing operates on GPU pointer tables and completes in constant time independent of $m$, an asymptotic improvement over the linear-in-$m$ cost of CPU-mediated serialization used by some block-paged designs.

The radix is admission-aware: prefix saves are gated against the cell budget of Section~3.7, and eviction follows a leaf-oldest policy that preserves long shared branches (system prompts, tool schemas) while shedding stale tails. Saves are incremental: when a new request extends a cached branch, only the delta cells are committed.

\subsubsection{Response and Render Caches}

Beyond prefix caching, the proposed architecture caches the \emph{full response text} for any prompt that has been served before. The sampler is seeded deterministically from an FNV-1a hash of the prompt tokens:

\begin{equation}
\text{seed}(P) = \mathrm{FNV1a}(P) \bmod 2^{32}
\end{equation}

so that identical prompts produce identical outputs at any temperature. A 1024-entry LRU response cache is consulted before the sequence pool is acquired: on hit, the request returns at network-roundtrip latency with zero GPU work. Two complementary caches handle the template-render and tokenize stages (256 and 64 entries respectively), eliminating redundant Jinja application and BPE tokenization, which on long prompts otherwise consume tens of milliseconds.

\section{Experimental Setup}

\subsection{Benchmark Scenario}

We designed the benchmark to simulate a realistic streaming workload where data arrives continuously and queries arrive sporadically after data updates:
\begin{enumerate}
    \item Initialize with 100 data samples
    \item For each of 15 iterations: add 55 new samples, then issue a query
    \item Measure query latency (time from query submission to response)
\end{enumerate}

This scenario specifically tests the streaming use case: data accumulates over time, and users query against the latest state. Context grows from 155 to 925 samples (approximately 2,600 to 14,800 tokens).

\subsection{Dataset}

We constructed a synthetic OHLCV dataset representing hourly candlestick bars for a single equity. Each sample contains open, high, low, close prices and volume ($\sim$16 tokens per sample). The data includes realistic market microstructure:
\begin{itemize}
    \item Uptrend phases with higher highs and higher lows
    \item Correction phases with pullbacks
    \item Consolidation phases with sideways movement
\end{itemize}

We report OHLCV results because financial streaming is the most latency-sensitive use case in our target workloads. The ingestion framework itself is data-type-agnostic: only the record schema and tokenisation rule are application-specific; the ring buffer, ingestion loop, Flash Queries, and caching infrastructure are shared.

\subsection{Query Set}

Six query types testing different analytical capabilities:
\begin{itemize}
    \item $Q_1$: Trend classification (UP/DOWN)
    \item $Q_2$: Pullback detection (YES/NO)
    \item $Q_3$: Recent high detection (YES/NO)
    \item $Q_4$: Consolidation detection (YES/NO)
    \item $Q_5$: Volume pattern detection (YES/NO)
    \item $Q_6$: Exact value retrieval (numeric)
\end{itemize}

$Q_6$ specifically tests precise information retrieval from mid-context, a known challenge for transformer models \cite{liu2024lost}.

\subsection{Baselines}

We compare against eight baselines spanning open-weight and proprietary models:
\begin{itemize}
    \item \textbf{vLLM}: High-throughput inference server with PagedAttention \cite{kwon2023vllm}
    \item \textbf{SGLang}: Structured generation framework with RadixAttention \cite{zheng2024sglang}
    \item \textbf{TensorRT-LLM}: NVIDIA's optimized inference library with FP8 quantization
    \item \textbf{llama.cpp}: Lightweight C++ inference engine with prompt caching enabled
    \item \textbf{OpenAI GPT-4o-mini}: Cloud API baseline (smaller, faster model)
    \item \textbf{OpenAI GPT-5.2}: Cloud API baseline (latest generation model)
    \item \textbf{Claude 3.5 Haiku}: Cloud API baseline (Anthropic's fast model)
    \item \textbf{Claude Opus 4.5}: Cloud API baseline (Anthropic's flagship model)
\end{itemize}

For cloud APIs, each query sends the complete accumulated context. While these services offer prompt caching for repeated prefixes, they do not maintain persistent conversational sessions across requests. All systems use temperature $\tau = 0$ (greedy decoding) for reproducibility.

\subsection{Hardware Configuration}

All open-weight model experiments (the proposed system, llama.cpp, vLLM, SGLang, and TensorRT-LLM) were conducted on a single NVIDIA L40S accelerator (48\,GB GDDR6, 864\,GB/s memory bandwidth) in a single-tenant configuration. All systems use Meta-Llama-3.1-8B-Instruct in BF16 precision ($\sim$16\,GB of model weights). The same model checkpoint, tokeniser, and chat template are used across every engine to isolate the contribution of the inference stack from any model-side variation.

\textbf{Speedup interpretation.} The $2.4\times$--$5.9\times$ speedup over conventional inference engines is an algorithmic improvement, not a hardware-specific one. Processing data at ingest time rather than query time reduces query latency complexity from $\mathcal{O}(n)$ to $\mathcal{O}(|q|)$ regardless of the accelerator. On more capable hardware the relative speedup should be preserved or amplified as larger contexts become feasible.

\textbf{Scaling projections.} On a single accelerator, the proposed architecture achieves sub-50\,ms median query latency ($\sim$43\,ms). Multi-accelerator deployments would enable larger concurrent session counts and potentially sub-20\,ms latency through parallelisation of the data plane.

\subsection{Metrics}

\begin{itemize}
    \item Query latency (ms): Time from query submission to response completion
    \item Accuracy (\%): Correct responses / total queries
    \item Latency scaling: How latency changes as context grows
\end{itemize}

\section{Results}

\begin{figure}[htbp]
\centering
\includegraphics[width=\textwidth]{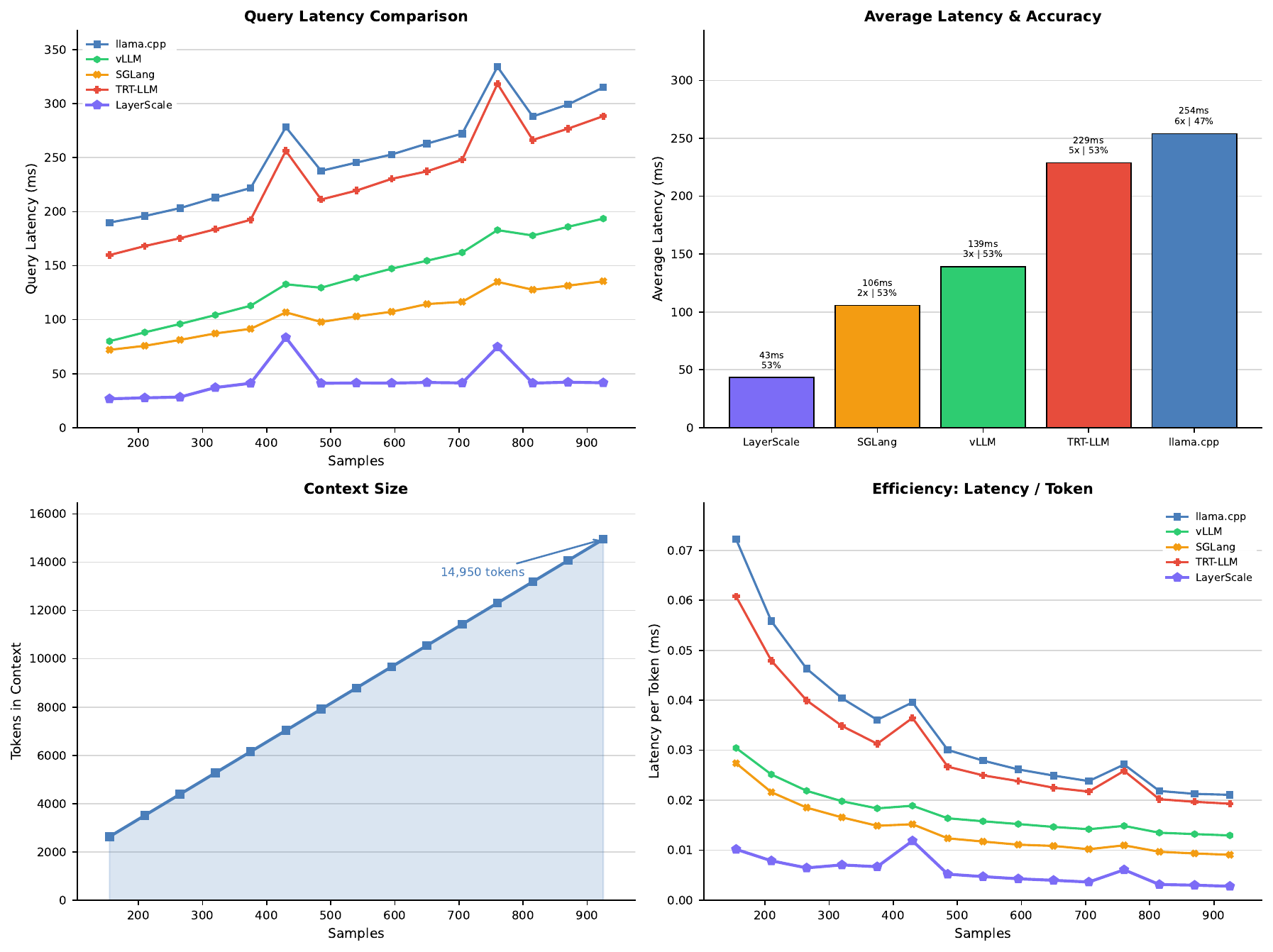}
\caption{Performance comparison across nine systems on streaming OHLCV data (155--925 samples). The proposed architecture maintains constant $\sim$43\,ms query latency regardless of context size. Conventional inference engines (vLLM, SGLang, TensorRT-LLM, llama.cpp) show $\sim$106--254\,ms latency, while cloud APIs range from 926\,ms to 3{,}962\,ms.}
\label{fig:latency-scaling}
\end{figure}

\begin{table}[htbp]
\caption{Performance comparison (streaming benchmark: 155--925 samples, 15 iterations)}
\label{tab:results}
\centering
\begin{tabular}{lrrr}
\toprule
Platform & Avg Latency (ms) & Accuracy (\%) & vs LayerScale \\
\midrule
LayerScale & 43.4 & 53.3 & baseline \\
SGLang & 105.6 & 53.3 & $2.4\times$ slower \\
vLLM & 139.1 & 53.3 & $3.2\times$ slower \\
TensorRT-LLM & 228.8 & 53.3 & $5.3\times$ slower \\
llama.cpp (with caching) & 254.0 & 46.7 & $5.9\times$ slower \\
GPT-5.2 & 926 & 75 & $21\times$ slower \\
GPT-4o-mini & 1{,}124 & 55 & $26\times$ slower \\
Claude Opus 4.5 & 2{,}628 & 70 & $61\times$ slower \\
Claude 3.5 Haiku & 3{,}962 & 65 & $92\times$ slower \\
\bottomrule
\end{tabular}
\end{table}

\subsection{Latency Analysis}

The proposed architecture shows \emph{constant} query latency regardless of context size. Conventional inference engines reach 106--254\,ms average latency on the same workload; the stateful path holds steady at $\sim$43\,ms throughout, a $2.4\times$--$5.9\times$ speedup.

\textbf{Why constant latency?} Under the access pattern characterised in Section~3, the query path scales with $|q|$ rather than with $n_{\text{ctx}}$: the dependence on accumulated context is collapsed into a constant-time term that does not grow with the streaming horizon. With short user queries ($|q| \approx 50$ in our benchmarks) against contexts that span thousands to tens of thousands of tokens, the observed query latency is dominated by the small per-query term and is therefore effectively independent of how much data has arrived since the session was opened. Single-token classification responses complete in tens of milliseconds.

\textbf{Inference engine comparison.} Despite their optimizations, request-driven inference engines must reprocess the full context every query:
\begin{itemize}
    \item SGLang reaches 105.6\,ms average latency with RadixAttention prefix caching
    \item vLLM reaches 139.1\,ms average latency with PagedAttention
    \item TensorRT-LLM reaches 228.8\,ms average latency
    \item llama.cpp reaches 254.0\,ms average latency with prompt caching enabled
\end{itemize}
The advantage on the streaming workload comes from amortising data ingestion across the inter-query interval, which all four request-driven systems cannot exploit because they discard intermediate state between requests. The gap to the closest competitor (SGLang) is the smallest because RadixAttention captures some incremental savings on a partially-shared context; the gap widens for the remaining systems.

\textbf{Cloud API comparison.} Cloud APIs must transmit and process the full context for each request:
\begin{itemize}
    \item GPT-5.2 and GPT-4o-mini reach $\sim$900--1{,}100\,ms average through optimized infrastructure
    \item Claude Opus 4.5 shows $\sim$2{,}600\,ms latency
    \item Claude 3.5 Haiku shows $\sim$4{,}000\,ms latency
\end{itemize}

The stateful path outperforms conventional inference engines by $2.4\times$--$5.9\times$ and cloud APIs by $21\times$--$92\times$.

\subsection{Latency Variance Analysis}

Beyond average latency, variance is critical for real-time applications. Table~\ref{tab:variance} presents standard deviation measurements across all systems.

\begin{table}[htbp]
\caption{Latency variance analysis (15 iterations)}
\label{tab:variance}
\centering
\begin{tabular}{lrrrr}
\toprule
Platform & Avg (ms) & Std Dev (ms) & Min (ms) & Max (ms) \\
\midrule
LayerScale & 43.4 & 15.6 & 26.8 & 83.4 \\
SGLang & 105.6 & 21.3 & 72.0 & 135.6 \\
vLLM & 139.1 & 37.2 & 80.1 & 193.6 \\
TensorRT-LLM & 228.8 & 47.5 & 159.8 & 318.4 \\
llama.cpp & 254.0 & 44.3 & 189.9 & 334.1 \\
GPT-5.2 & 926 & 213 & 697 & 1{,}590 \\
GPT-4o-mini & 1{,}124 & 423 & 602 & 2{,}209 \\
Claude Opus 4.5 & 2{,}628 & 261 & 2{,}215 & 3{,}251 \\
Claude 3.5 Haiku & 3{,}962 & 1{,}519 & 1{,}550 & 6{,}824 \\
\bottomrule
\end{tabular}
\end{table}

\textbf{Key observations:}
\begin{itemize}
    \item The stateful path exhibits the lowest absolute variance ($\sigma = 15.6$\,ms)
    \item Open-weight inference engines show moderate variance ($\sigma = 21$--$48$\,ms)
    \item Cloud APIs exhibit high variance, particularly Claude 3.5 Haiku ($\sigma = 1{,}519$\,ms) due to network latency and shared infrastructure
\end{itemize}

For latency-sensitive applications, low variance is as important as low mean latency: the proposed architecture provides both.

\subsection{Accuracy}

All four open-weight engines run the same Meta-Llama-3.1-8B-Instruct checkpoint and reach the same 53.3\% accuracy on the streaming query set, with the exception of llama.cpp at 46.7\%. Cloud models (GPT-5.2 75\%, Claude Opus 4.5 70\%, Claude 3.5 Haiku 65\%, GPT-4o-mini 55\%) generally score higher on the harder reasoning queries, reflecting model scale rather than serving architecture. The architectural optimizations described in Section~3 preserve model quality while delivering substantially lower latency.

\section{Discussion}

\subsection{Applicability}

Our approach is applicable when:
\begin{enumerate}
    \item[(i)] Context has static/dynamic partitioning
    \item[(ii)] Data arrives continuously but queries arrive sporadically
    \item[(iii)] Data updates are incremental (append or sliding window)
    \item[(iv)] Low query latency is required
\end{enumerate}

Financial streaming is a natural fit. Other applicable domains include log analysis, sensor monitoring, IoT data processing, and conversational agents with persistent memory.

\subsection{Limitations}

The current implementation assumes:
\begin{itemize}
    \item Single-sequence contexts (no cross-document attention)
    \item Causal attention patterns
    \item Sufficient memory to maintain session state
\end{itemize}

Multi-document retrieval and bidirectional models require architectural extensions not addressed in this work.

\textbf{Attention scope.} Within the sliding window, we maintain full quadratic attention over all retained data. However, when data is evicted from Region 1, attention over that data is lost. The model retains only what remains in the active context. For applications requiring awareness of historical patterns beyond the retention window, a promising architectural extension is to add a dedicated KV cache region for rolling summaries: periodically compressing older data into summary representations before eviction, preserving semantic information while bounding memory growth.

\subsection{Memory Considerations}
\label{sec:memory}

We trade memory for computation. Each session maintains intermediate representations proportional to context length. The KV cache memory footprint per session is:
\begin{equation}
M_{\text{KV}} = 2 \cdot L \cdot d \cdot n_{\text{ctx}} \cdot \text{sizeof}(\text{dtype})
\label{eq:memory}
\end{equation}
where $L$ is the number of layers, $d$ is the hidden dimension, $n_{\text{ctx}}$ is the context length, and the factor of 2 accounts for both keys and values. For Meta-Llama-3.1-8B ($L = 32$, $d = 4096$) with 32K context and FP16 precision:
\begin{equation}
M_{\text{KV}} = 2 \cdot 32 \cdot 4096 \cdot 32768 \cdot 2 \text{ bytes} \approx 17 \text{ GB per session}
\end{equation}

For deployments with many concurrent sessions, memory management matters. The maximum concurrent sessions is bounded by:
\begin{equation}
N_{\text{sessions}} \leq \frac{M_{\text{available}} - M_{\text{model}}}{M_{\text{KV}}}
\label{eq:sessions}
\end{equation}
We implement session pooling and LRU eviction for production deployments, prioritizing active sessions while evicting idle ones to disk or discarding them entirely.

\section{Related Work}

\textbf{KV-Cache Serving Frameworks.} Production inference systems like vLLM \cite{kwon2023vllm}, TensorRT-LLM, SGLang \cite{zheng2024sglang}, llama.cpp, Ollama, and HuggingFace TGI implement KV-cache persistence and prefix sharing to amortize computation across requests with common prefixes. These systems are \emph{request-driven}: computation occurs when queries arrive, and caching optimizes for repeated identical prefixes across multiple users. We address a different problem: \emph{data-driven} inference where a single session receives continuous data updates and sporadic queries. The optimization target shifts from ``serve many users with shared static prefixes'' to ``serve one session with continuously evolving context.''

\textbf{Prefix Caching.} RadixAttention (SGLang) and automatic prefix caching (vLLM) identify and reuse common prefixes across requests. These techniques assume prefixes are static and shared. In streaming workloads, the ``prefix'' changes with each data update, invalidating cached state. Our region-based architecture treats the dynamic portion as first-class, processing updates incrementally rather than invalidating on change. In companion work, we extend prefix caching with response-level caching and n-gram speculative generation for multi-agent tool calling workloads, where conversation prefixes grow monotonically across turns.

\textbf{Disaggregated Prefill.} vLLM's disaggregated prefilling \cite{zhong2024distserve} separates prompt processing (prefill) from token generation (decode) across different instances, transferring KV caches via connectors. This improves time-to-first-token by preventing prefill jobs from blocking decode operations. However, it remains request-driven: prefill still occurs when queries arrive, and KV cache must be transferred between instances for each request. The approach optimizes latency distribution but explicitly does not improve throughput. Our approach is fundamentally different: we process data into the KV cache \emph{before} any query arrives, eliminating prefill from the query path entirely. There is no KV cache transfer overhead because the cache already resides where queries execute. For streaming workloads, this yields both lower latency and higher throughput.

\textbf{Prompt Caching.} Cloud providers \cite{anthropic2024prompt} offer prompt caching for repeated identical prompts. This fails for streaming: each query has a different prompt (new data appended), so cache hit rate approaches zero.

\textbf{Sparse Attention.} Longformer \cite{beltagy2020longformer}, BigBird \cite{zaheer2020bigbird}, and Sparse Transformer \cite{child2019sparsetransformer} reduce attention complexity from $\mathcal{O}(n^2)$ to $\mathcal{O}(n)$ by limiting which tokens can attend to which: sliding windows, random patterns, or strided access. The drawback: sparse patterns sacrifice full global context access. Long-range dependencies may be missed if they don't align with the sparsity pattern, degrading performance on tasks requiring precise information retrieval across the full context. We maintain full quadratic attention while achieving constant query latency through architectural separation of data and query processing.

\textbf{FlashAttention.} FlashAttention \cite{dao2022flashattention} and FlashAttention-2 \cite{dao2023flashattention2} optimize the \emph{implementation} of attention by exploiting GPU memory hierarchy: tiling attention computation to maximize SRAM utilization and minimize HBM transfers. This achieves 2--4$\times$ wall-clock speedups while computing mathematically identical results. FlashAttention is orthogonal to our contribution: it accelerates individual attention operations, while we eliminate repeated attention operations through architectural redesign. Our system can use FlashAttention as the underlying attention kernel while still benefiting from data-driven processing that avoids redundant computation.

\textbf{RAG.} Retrieval-augmented generation \cite{lewis2020retrieval} externalizes context to a retrieval system. This introduces retrieval latency, recall failures, and loses the benefits of full attention over complete data. We keep all data in-context with full attention, avoiding retrieval overhead and ensuring complete context access.

\textbf{Summarization Chains.} Approaches that summarize context to fit within token limits lose information and introduce summarization errors. We maintain full fidelity of the original data without lossy compression.

\textbf{Recursive Language Models.} Recent work on Recursive Language Models (RLMs) \cite{zhang2025rlm} addresses long-context processing by allowing models to programmatically decompose and recursively call themselves over context snippets. RLMs can process inputs up to 100$\times$ beyond the native context window by treating long prompts as an external environment. However, this approach has significant latency implications: each recursive call is blocking with no prefix caching, resulting in query latencies ranging from ``a few seconds to several minutes'' depending on decomposition strategy \cite{zhang2025rlm}. The framework explicitly ``did not optimize implementation for speed'' and provides ``no strong guarantees about controlling either the total API cost or the total runtime.''

Our approach differs fundamentally in optimization target. RLMs optimize for \emph{context capacity}: handling arbitrarily long inputs at the cost of unpredictable, high latency. We optimize for \emph{query latency}: achieving constant tens-of-milliseconds responses within the model's context window. For streaming applications requiring real-time responses, the difference is stark: RLMs may take minutes per query while the proposed architecture maintains sub-50\,ms latency regardless of accumulated context. Additionally, RLMs require multiple recursive LLM calls with information potentially lost during decomposition, while we preserve full quadratic attention over all retained data in a single forward pass. The approaches are complementary: RLMs excel at one-shot processing of very long documents; we excel at continuous streaming with sporadic low-latency queries.

\section{Conclusion}

Request-driven inference frameworks optimize for the wrong workload when applied to streaming applications. Prefix caching assumes static, shared prefixes; streaming workloads have dynamic, per-session context. Cache hit rates approach zero, and every query pays full $\mathcal{O}(n)$ cost.

We exploit the asymmetry between data and query arrival rates and move prefill off the critical path: data arrival triggers processing, not queries. This enables $\mathcal{O}(|q|)$ query latency regardless of accumulated context size.

Results on streaming benchmarks:
\begin{itemize}
    \item \textbf{Sub-50\,ms query latency}: $\sim$43\,ms average as context grows from 2{,}600 to 14{,}800 tokens, while request-driven inference engines reach 106--254\,ms on the same workload
    \item \textbf{$2.4\times$--$5.9\times$ speedup} over conventional inference engines (vLLM, SGLang, TensorRT-LLM, llama.cpp)
    \item \textbf{$21\times$--$92\times$ speedup} over cloud APIs (GPT-4o-mini, GPT-5.2, Claude Haiku, Claude Opus 4.5)
    \item \textbf{Continuous ingestion}: background processing enables true streaming with non-blocking data push, and only the new delta is ever processed
    \item \textbf{Full attention over retained context}: unlike sparse attention approaches, we maintain complete quadratic attention over all data within the retention window
\end{itemize}

Flash Queries extend this further: by pre-evaluating registered questions against evolving context state during idle GPU cycles, we push latency from $\sim$43\,ms toward $\sim$2\,ms for predictable query patterns. Combined with ready-position speculative exit, the system can return cached or high-confidence answers with zero GPU work at query time. For workloads where 60--80\% of queries are predictable, the effective average latency drops to single-digit milliseconds, approaching the theoretical minimum of network round-trip time.

The contribution is architectural: recognising that streaming workloads require data-driven, not request-driven, computation models. Existing KV-cache techniques are necessary but not sufficient. We provide the session management, region semantics, incremental update mechanisms, and ahead-of-time query evaluation that streaming applications require.

A second consequence, with broader operational impact than per-query latency, is GPU utilisation. Large inference deployments typically run at sustained accelerator utilisation in the 10--45\% band, with the gap to peak dominated by between-request idle time on accelerators that have nothing useful to do. The architecture described here turns that idle time into pre-computed Flash Query answers against an always-current KV cache, so the same hardware delivers a higher rate of useful analytical output per unit time and per unit power, without any change to model weights or kernel implementations. For operators of large clusters, this is the rare class of optimization that simultaneously lowers latency, raises throughput, and improves accelerator utilisation.

\bibliographystyle{plainnat}

\end{document}